\documentclass[conference]{IEEEtran}
\ifCLASSINFOpdf
\else
\fi

\usepackage{mathtools}
\usepackage{amsmath}
\usepackage{amssymb}
\usepackage{graphicx}
\usepackage[table]{xcolor}
\usepackage{adjustbox}

\usepackage{pifont}
\newcommand{\cmark}{\ding{52}}%
\newcommand{\xmark}{\ding{56}}%

\usepackage{tikz}
\usetikzlibrary{positioning}
\newdimen\nodeDist
\usepackage[lined,linesnumbered,ruled,commentsnumbered]{algorithm2e}

\newcommand\R{{\mathbb R}}

\newcommand\N{{\mathbb N}}
\usepackage{amsmath}

\DeclareMathOperator*\argmin{\operatorname{argmin}}

\newcommand{\BigO}[1]{\ensuremath{\operatorname{ \mathcal{O}}\left(#1\right)}}

\usepackage[english]{babel}
\usepackage{blindtext}

\usepackage{url} 
\urldef{\mails}\path|{josif, wistuba,  schmidt-thieme}@ismll.uni-hildesheim.de|

\begin{document}
%
\title{Scalable Discovery of Time-Series Shapelets}

\author{\IEEEauthorblockN{Josif Grabocka, Martin Wistuba, Lars Schmidt-Thieme}
\IEEEauthorblockA{Information Systems and Machine Learning Lab\\
University of Hildesheim, 31141 Hildesheim, Germany\\
Email: \{josif,wistuba,schidt-thieme\}@ismll.uni-hildesheim.de}
}


%


\maketitle

\begin{abstract}
Time-series classification is an important problem for the data mining community due to the wide range of application domains involving time-series data. A recent paradigm, called shapelets, represents patterns that are highly predictive for the target variable. Shapelets are discovered by measuring the prediction accuracy of a set of potential (shapelet) candidates. The candidates typically consist of all the segments of a dataset, therefore, the discovery of shapelets is computationally expensive. This paper proposes a novel method that avoids measuring the prediction accuracy of similar candidates in Euclidean distance space, through an online clustering pruning technique. In addition, our algorithm incorporates a supervised shapelet selection that filters out only those candidates that improve classification accuracy. Empirical evidence on 45 datasets from the UCR collection demonstrate that our method is 3-4 orders of magnitudes faster than the fastest existing shapelet-discovery method, while providing better prediction accuracy. 
\end{abstract}


%
\IEEEpeerreviewmaketitle

\section{Introduction}

Classification of time-series data has attracted considerable interest in the recent decades, which is not surprising given the numerous domains where time series are collected. A recent paradigm has emerged into the perspective of classifying time series, the notion of \emph{shapelets}. Shapelets are supervised segments of series that are highly descriptive of the target variable~\cite{ye2009time}. In the recent years, shapelets have achieved a high momentum in terms of research focus \cite{ye2009time,lines2012shapelet,zakaria2012clustering,mueen2011logical,rakthanmanon2013fast}.

Distances of time series to shapelets can be perceived as new classification predictors, also baptized as  \emph{"the shapelet-transformed data"} \cite{lines2012shapelet,hills2013transform}. It has been shown by various researchers that shapelet-derived predictors boost the classification accuracy \cite{ye2011time,lines2012shapelet,mueen2011logical}. In particular, shapelets are efficient in datasets where the class discrimination is attributed to local variations of the series content, instead of the global structure \cite{ye2009time}. Even though not explicitly mentioned by the related work, the discovery of shapelets can be categorized as a supervised dimensionality reduction technique. In addition, shapelets also provide interpretive features that help domain experts understand the differences between the target classes.

The discovery of shapelets, on the other hand, has not been as enthusiastic as their prediction accuracy. The current discovery methods need to search for the most predictive shapelets from all the possible segments of a time series dataset \cite{ye2009time,mueen2011logical,lines2012shapelet}. Since the number of possible candidates is high, the required time for evaluating the prediction quality of each candidate is prohibitive for large datasets. Therefore, the time series research community has proposed different speed-up techniques \cite{ye2009time,mueen2011logical,rakthanmanon2013fast}, aiming at making shapelet discovery \textbf{feasible} in terms of time.

This paper proposes a novel method that discovers time-series shapelets considerably faster than the fastest existing method. Our method follows the knowledge that time-series instances contain lots of similar segments. Often inter-class variations of time series depend on differences within small segments, with the remaining parts of the series being similar. Therefore, we hypothesize that the time needed to discover shapelets can be \textbf{scaled}-up by pruning candidate segments that are similar in Euclidean distance space. We introduce a fast distance-based clustering approach to prune future segments that result similar to previously considered ones. In addition, we propose a fast supervised selection of shapelets that filters out the qualitative shapelets using an incremental nearest-neighbor classifier. Extensive experiments conducted on real-life data demonstrate a large reduction (3-4 orders of magnitude) of the discovery time, by even gaining prediction accuracy with respect to baselines. The contributions of this paper can be short-listed as follows:

\begin{enumerate}
\item A fast pruning strategy for similar shapelets in Euclidean space involving a distance-based clustering approach;
\item A fast supervised selection of qualitative shapelets using an incremental nearest-neighbor classifier, conducted jointly with the pruning;
\item Extensive experimental results against the fastest existing shapelet discovery methods on a large set of 45 time-series datasets.
\end{enumerate}



\section{Related Work}

Shapelets were introduced by \cite{ye2009time} as a new primitive representation of time-series that is highly predictive of the target. A large pool of candidates from all segments of a dataset were assessed as potential shapelet candidates, while the minimum distance of series to shapelets was used as a predictive feature. The best performing candidates were ranked using the information gain criteria over the target. Successively, other prediction quality metrics were also elaborated such as the Kruskal-Wallis or Mood's median \cite{hills2013transform}, as well as F-Stats \cite{lines12quality}. The minimum distance of the time-series to a set of shapelets can be categorized as a data transformation (dimensionality reduction) and is named as \emph{shapelet-transformed data} \cite{lines2012shapelet}. Standard classifiers have been shown to perform competitively over the shapelet-transformed predictors \cite{hills2013transform}.

\vspace{-0.1cm}

\begin{figure*}[]
\centering
\includegraphics[scale=0.67, trim=2cm 15.00cm 0.0cm 7.0cm]{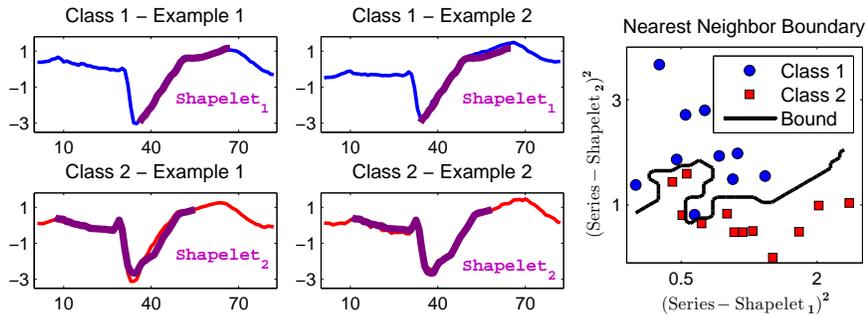}
\caption[]{TwoLeadECG dataset: Aligning shapelets to the closest series segments, and on right the resulting 2-dimensional shapelet-transformed training data.}
\label{fig:shapeletTransform}
\end{figure*}

The excessive amount of potential candidates makes the brute-force (exhaustive) shapelet discovery intractable for large datasets. Therefore, researchers have come up with various approaches for speeding up the search. Early abandoning of the Euclidean distance computation combined with an entropy pruning of the information gain metric is an early pioneer in that context \cite{ye2009time}. Additional papers emphasize the reuse of computations and the pruning of the search space \cite{mueen2011logical}, while the projection of series to the SAX representation was also elaborated \cite{rakthanmanon2013fast}. Furthermore, the discovery time of shapelets has been minimized by mining infrequent shapelet candidates \cite{he2012fast}. Speed-ups have also been attempted by using hardware-based implementations, such as the usage of the processing power of GPUs for boosting the search time \cite{chang2012efficient}.

In terms of applicability, shapelets have been utilized in a battery of real-life domains. Unsupervised shapelets discovery, for instance, has been shown useful in clustering time series \cite{zakaria2012clustering}. Shapelets have seen action in classifying/identifying humans through their gait patterns \cite{sivakumar2012human}. Gesture recognition is another application domain where the discovery of shapelets has played an instrumental role in improving the prediction accuracy \cite{hartmann2010gesture,hartmann2010prototype}. In the realm of medical and health informatics, interpretable shapelets have been shown to help the early classification of time-series~\cite{xing2011extracting,xing2012early}.

In comparison to the state-or-the-art methods, we propose a fast novel method that discovers shapelets by combining a pruning strategy of similar candidates with an incremental classification technique.

\section{Scalable Shapelet Discovery}

\subsection{Distances of Shapelets to Series as Classification Features}

Throughout this paper we denote a time-series dataset having $N$ series of $M$ points each, as $T \in \R^{N \times M}$. While our method can work with series of arbitrary lengths, we define a single length M for ease of mathematical formalism. The distances of shapelets to series can be used as classification features, also known as shapelet-transformed features \cite{lines2012shapelet,hills2013transform}. The distance of a candidate shapelet to the closest segment of a series can be perceived as a membership degree for that particular shapelet. Equations~\ref{eq:minDistances1} and \ref{eq:minDistances2} formalize the minimum distances between a shapelet $s \in \R^m$ and the dataset $T$ as a vector of the Euclidean distances ($\mathcal{D}$) between the shapelet and the closest segment of each series. (The notation $V_{a:b}$ denotes a sub-sequence of vector $V$ from the $a$-th element to the $b$-th element.)

%


\begin{eqnarray}
\label{eq:minDistances1}
\textbf{MinDist}(s,T) &:=& \begin{bmatrix}
       \mathcal{D}(s,T_1) \\
       \mathcal{D}(s,T_2) \\
       \vdots \\
       \mathcal{D}(s,T_N) 
     \end{bmatrix} 
\\
\mathcal{D}(s,T_i) &:=& \min_{j = 1,\dots,M-m+1} \left\| T_{i,j:j+m-1} - s \right\|^2
\label{eq:minDistances2}
\end{eqnarray}

An illustration of the minimum distances between shapelets and series is shown in Figure~\ref{fig:shapeletTransform} for the TwoLeadECG dataset. Two shapelets (purple) are matched to four time-series of two different classes (red and blue). Following the principle that Equation~\ref{eq:minDistances2} states, the distance of a shapelet is computed to the closest series segment. The distances between training time-series and the two shapelets can project the dataset to a 2-dimensional shapelet-transformed space, as shown on the right sub-plot. A nearest neighbor classifier and the corresponding classification decision boundary is also illustrated.

\subsection{Quantification of Similarity Using a Distance Threshold}

A time series dataset contains lots of similar patterns spread over various instances. Since series from the same class follow a similar structure, similar patterns repeat over time-series of the same class. Similarities can also be observed among time series of different classes, because often classes are discriminated by differences in small sub-sequences rather than the global structure. As a result, we raise the hypothesis that existing state-of-the-art techniques, which exhaustively search all candidates, inefficiently consider lots of very similar patterns.

\begin{figure}[h]
\centering
\includegraphics[scale=0.5, trim=2.0cm 15cm 0.0cm 7cm]{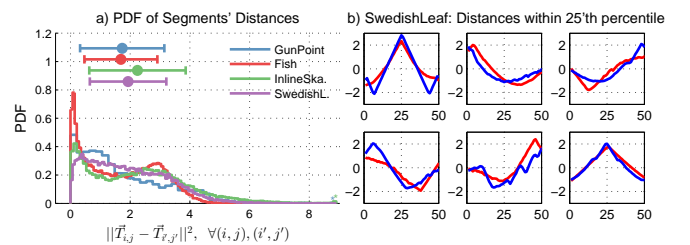}
\caption[]{\textbf{a)}Distribution of distances among random pairs of candidates; \textbf{b)} Illustration of similar segments from the SwedishLeaf dataset with pairwise distances less than the $25-$th percentile of the distribution in \textbf{a)}.}
\label{fig:epsilonThreshold}
\end{figure}

Figure~\ref{fig:epsilonThreshold} illustrates the distribution of distances among arbitrary pairs of candidate segments from various time series of the UCR collection of datasets~\cite{ucrDatasets}. As can be seen from sub-figure \emph{a)}, the distribution of distances is highly skewed towards zero, which indicate that most candidates are very similar to each other. However, a threshold separation on the similarity distance is required to judge segments as being similar or not. We propose to use a threshold over the percentile on the distribution of distances. For instance, Figure \emph{\ref{fig:epsilonThreshold}.b)} displays pairs of similar segments whose pairwise distances are within the 25-th percentile of the distance distribution.

\begin{algorithm}[!]
 \KwData{Time series data $T \in \R^{N \times M}$, Percentile $p \in \left[ 1,\dots,100 \right]$, Shapelet Lengths $\Phi \in \N^L$} 
 \KwResult{ Threshold distance $\epsilon \in \R$ } 
 $Z \leftarrow \emptyset$\; 
 \For{$1,\dots,NM$}{ 
 	Draw random shapelet length $\Phi_l \sim  \mathcal{U}(\Phi_1,\dots,\Phi_L)$ \; 
 	Draw segment indices $(i,j) \sim \left( \mathcal{U}(1,\dots,N), \mathcal{U}(1,\dots,M-\Phi_l+1) \right)$ \; 
 	Draw segment indices $(i',j') \sim \left( \mathcal{U}(1,\dots,N), \mathcal{U}(1,\dots,M-\Phi_l+1) \right)$ \; 	 
 	$Z \leftarrow Z \cup \left\{ \frac{1}{\Phi_l}  || T_{i,j:j+\Phi_l-1} - T_{i',j':j'+\Phi_l-1}||^2 \right\} $\; 
 } 
 $Z \leftarrow \mbox{sort}(Z)$\; 
 $\epsilon \leftarrow Z_{ \lceil \frac{p}{100} N \, M \rceil }$\;  
 \KwRet{$ \epsilon$} 
 \caption{ \textbf{ComputeThreshold}: Compute the pruning similarity distance threshold $\epsilon$.} 
  \label{alg:computeThreshold}  
\end{algorithm}

The procedure of determining a distance threshold value, denoted $\epsilon$ and belonging to the $p$-th percentile of the distance distribution, is described in Algorithm~\ref{alg:computeThreshold}. The algorithm selects a pair of random segments starting at indices $(i,j), (i',j')$ and having random shapelet lengths $\Phi_l$. Then a distribution is build by accumulating the distances of random pairs of segments and the distance value that corresponds to the desired percentile $p$ is computed from the sorted list of distance values. For instance, in case all the distance values are sorted from smallest to largest, then the 25-th percentile is the value at the index that belongs to $25\%$ of the total indices.

Totally there are $N M L$ segments in a time-series dataset and the total number of pairs is $\frac{1}{2} (NML)(NML-1)$. However, in order to guess the distribution of a set of values (here distances), one doesn't need to have access to the full population of values. On the contrary, a sample of values are sufficient for estimating the distribution. In order to balance between a fast and accurate compromise we choose to select $NM$-many random segment pairs for estimating the distance distributions. The runtime speed up success of Section~\ref{resultsSec} indicates that the distance threshold estimation is accurate.

\subsection{Main Method: Scalable Discovery of Time-series Shapelets} 

The scalable discovery of time series shapelets follows the two primary principles of this paper: \emph{i)} Pruning of similar candidates, and \emph{ii)} on-the-fly supervised selection of shapelets. The rationale of these principles is based on the knowledge that the majority of patterns from any specific time series are similar to patterns in other series of the same dataset. Therefore, it is computationally non-optimal to measure the quality of lots of very similar candidates. Instead, we aim at considering only a small nucleus of non-redundant candidates. 

\begin{algorithm}[!t]
 \KwData{Time series data $T \in \R^{N \times M}$, Labels $Y \in \R^{N}$ Distance Threshold Percentile $p \in \left[1,\dots,100\right]$, Piecewise Aggregate Approximation ratio: $r \in \{\frac{1}{2}, \frac{1}{4}, \dots \}$, Shapelet lengths: $\Phi \in \N^L $ }
 \KwResult{Accepted shapelets list $\mathcal{A} \in \R^{* \times *}$, Minimum Distances $D \in \R^{* \times *}$ }
 $\epsilon \leftarrow \mbox{ComputeThreshold}(T,p,\Phi) $\;
 $\mathcal{A}~\leftarrow~\emptyset,  \mathcal{R}~\leftarrow~\emptyset,  D~\leftarrow~\emptyset,  X~\leftarrow~\textbf{0}_{N \times N}, \mbox{prevAccuracy} \leftarrow - \infty$\;
 \For{$1,\dots, N M L$}{ 
 Draw random series: $i \sim \mathcal{U}\{1,\dots,N\} $\;
 Draw random shapelet length: $\Phi_l \sim \mathcal{U}\{\Phi_1,\dots,\Phi_L\} $\;
 Draw random segment start: $j \sim  \mathcal{U}\{1,\dots,M-\Phi_l+1\} $\;
 Selected random candidate: $s \leftarrow T_{i,j:j+\Phi_l-1}$\;
	\If{ $\lnot$LookUp($s,\mathcal{A},\epsilon$) $\land$ $\lnot$LookUp($s,\mathcal{R},\epsilon$)}{
		$d^s \leftarrow$ MinDist($s,T$)  \;
		\For{$i = 1,\dots,N; \; m = i+1,\dots,N$ }{
					$X_{i,m} \leftarrow X_{i,m} {\bf +} \left(d^s_i - d^s_m\right)^2$\;
					}
		\eIf{ $\mbox{Accuracy}(X, Y) > \mbox{prevAccuracy}$   }{
				$\mathcal{A} \leftarrow \mathcal{A} \cup \{ s \}$\;
				$D \leftarrow D \cup \{ d^s \}$\;
				$\mbox{prevAccuracy} \leftarrow \mbox{Accuracy}(X, Y)$\;
			}
		{ 
			$\mathcal{R} \leftarrow \mathcal{R} \cup \{ s \}$\;
			\For{$i = 1,\dots,N; \; m = i+1,\dots,N$ }{
			$X_{i,m} \leftarrow X_{i,m} {\bf -} \left(d^s_i - d^s_m\right)^2 $\;
			}
		}
	}
 }
 \KwRet{$\mathcal{A},D$}
 \caption{ \textbf{DiscoverShapelets}: Scalable discovery of shapelets }
 \label{alg:discoverShapelets}
\end{algorithm}

\subsubsection{Taxonomy of The Terms} By \textbf{refused} candidates we mean the candidates that are similar to previously considered ones, while by \textbf{considered} candidates we mean those who are not refused. Among the \textbf{considered} candidates, some of them will be \textbf{accepted} and the rest \textbf{rejected}. The decision tree below helps clarifying the terms.

\begin{tikzpicture}[
	thick,
	scale=0.85,
	node/.style={ %
	transform shape
	},
]

\node [node] (A) {Is \emph{candidate} similar to previously considered ones?};
\path (A) ++(-150:25mm) node [node] (B) {\textbf{REFUSE} \emph{candidate}!};
\path (A) ++(-31:\nodeDist) node [node] (C) {Does \emph{candidate} improve accuracy?};
\path (C) ++(-150:\nodeDist) node [node] (D) {\textbf{ACCEPT} \emph{candidate}!};
\path (C) ++(-30:\nodeDist) node [node] (E) {\textbf{REJECT} \emph{candidate}!};

\draw (A) -- (B) node [left,pos=0.35] {Yes.} (A);

\draw (A) -- (C) node [right,pos=0.25] {{\footnotesize No. Then \textbf{CONSIDER} \emph{candidate}!}} (C);

\draw (C) -- (D) node [left,pos=0.25] {Yes.} (D);

\draw (C) -- (E) node [right,pos=0.25] {No.} (E);

\end{tikzpicture}

The similarity of a candidate is first evaluated by looking up whether a close candidate has been previously considered, i.e has been previously flagged as either accepted or rejected. The considered non-redundant (non-similar to previous) candidates are subsequently checked on whether they improve the classification accuracy of previously selected candidates, and are either marked as accepted or rejected. 

We are presenting our method as Algorithm~\ref{alg:discoverShapelets} and incrementally walking the reader through the steps. The algorithm is started by compressing the time-series via the Piecewise Aggregate Approximation technique, to be detailed in Section~\ref{sec:paa}. In order to prune similar candidates, the threshold distance $\epsilon$ is computed using Algorithm~\ref{alg:computeThreshold}. Our method operates by populating two lists of accepted and rejected shapelets, denoted as $\mathcal{A}$ and $\mathcal{R}$, and storing a distance matrix $X$ for distances between series in the shapelet-transformed space.

\begin{figure*}[]
\centering
\includegraphics[scale=0.69, trim=1.5cm 12.00cm 0.0cm 7.0cm]{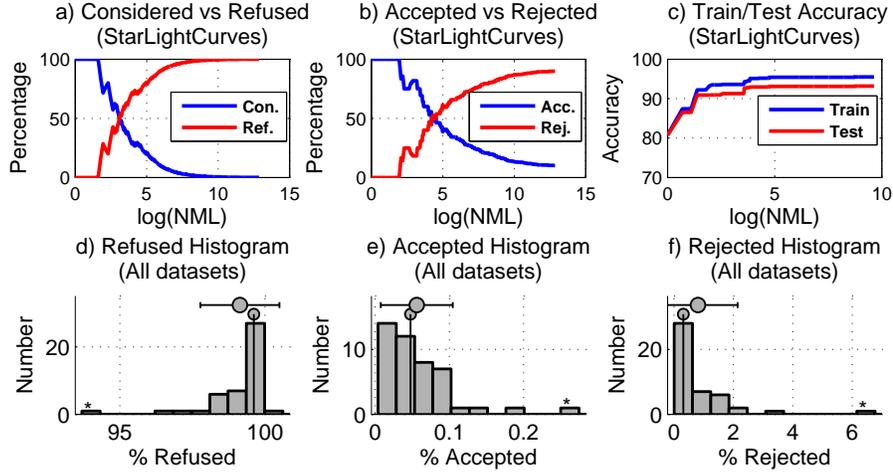}
\caption[]{ \textbf{a,b,c)} Relations of refused, rejected and accepted candidate shapelets, and the resulting accuracy, for the Starlight dataset; \textbf{d,e,f)} Histograms of refused, accepted and rejected candidate percentages over all 45 UCR datasets.}
\label{fig:accRejRef}
\end{figure*}

\subsubsection{Pruning Similar Candidates} Random shapelet candidates, denoted $s$, are drawn from the training time-series and a similarity search is conducted by looking up whether similar candidates have been previously considered (lines 4-8). Equation~\ref{eq:lookUp} formalizes the procedure as a similarity search over a list $\mathcal{L}$ (e.g., $\mathcal{A}$ or $\mathcal{R}$), considering candidates having same length ($len()$). Please note that in the concrete implementation we use a pruning of the Euclidean distance computations, by stopping comparisons exceeding the threshold $\epsilon$. 
\begin{eqnarray}
\nonumber
\textbf{LookUp}(s, \mathcal{L},\epsilon) \;\; := \;\;  \exists q \in \mathcal{L} \;\;\; | \;\;\;  || s - q ||^2 < \epsilon \\ \land \, \text{length}(s) = \text{length}(q)
\label{eq:lookUp}
\end{eqnarray}

\vspace{-0.1cm}

In case a candidate is found to be novel (not similar to previously considered), then the distance of the candidate to training series are computed using Equation~\ref{eq:minDistances1} and stored as $d^s$. Our approach evaluates the \textbf{joint} accuracy of accepted shapelets, so far, using a nearest neighbor classifier over the shapelet-transformed data, i.e. distances of series to accepted shapelets. When checking how does a new $(n+1)$-st candidate influence the accuracy of $n$ currently accepted candidates, an important speed-up trick can be used. We can pre-compute the distances among shapelet-transformed features in an incremental fashion. The distances among series in the feature-transformed space are stored in a distance matrix, denoted $X$, and the contribution of a new candidate can be simply added to the distance matrix. Those steps correspond to lines 10-12 and 19-21 in Algorithm~\ref{alg:discoverShapelets}. It is trivial to verify that this technique can improve the run-time of a nearest neighbor from $\BigO{N^2 |\mathcal{A}|}$ to $\BigO{N^2}$, which means that we can avoid recomputing distances among previously accepted $|\mathcal{A}|$-many shapelets.

\begin{figure*}[!htbp] 
\centering
\includegraphics[scale=0.7, trim=1.5cm 15.0cm 0.0cm 7.0cm]{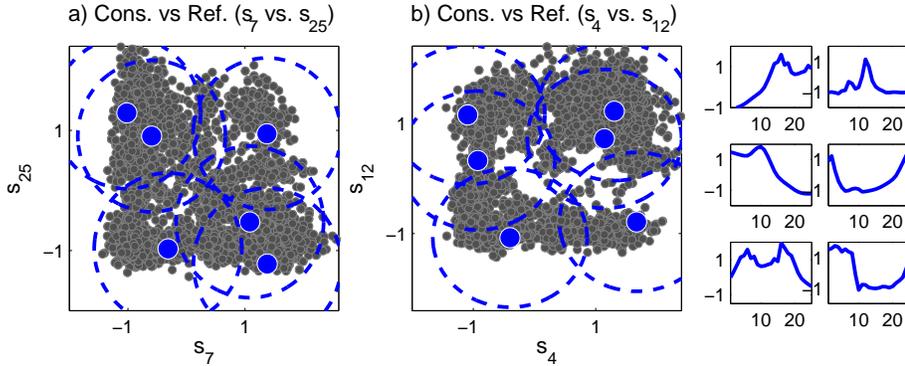}
\caption[]{Refused (gray) candidates versus Considered ones (blue), together with the distance threshold circles, are shown for MALLAT dataset. Considered shapelets are displayed on the right. Parameters: $r=0.125$, $p=25$, (i.e. radius is $\epsilon=1.26$), $m=25$. }
\label{fig:clusterIllustration}
\end{figure*}

\subsubsection{Supervised Shapelet Selection} In case the contribution of a unique candidate improves the classification accuracy of a nearest neighbor classifier, then the shapelet is added to the accepted list and the distance vector is stored in a shapelet-transformed data representation $\mathcal{D}$, in order to be later on used for classifying the test instances. Otherwise, the shapelet is inserted to the rejected list and the contribution of the candidate to the distance matrix $X$ is rolled back. The classification accuracy of the distances between series and a set of shapelets is measured by the nearest neighbor accuracy of the cumulative distance matrix $X$. The accuracy over the training data is formalized in Equation~\ref{eq:nearestNeighbor}.

\vspace{-0.2cm}
\begin{eqnarray}
\textbf{Accuracy}(X,Y) := \frac{1}{N} \left| \left\{ i \; | \;\; Y_i = Y_{ \argmin_{m, m \ne i } X_{i,m} } \right\}  \right|_{i=1}^{N}
\label{eq:nearestNeighbor} 
\end{eqnarray}

\subsubsection{Number of Sampled Candidates}

Algorithm~\ref{alg:discoverShapelets} samples shapelet candidates randomly, however the total number of sampled candidates is $NML$, that upper bounds the total possible series segments of a dataset. Our method could perform competitively even if we would sample a subset of the total possible candidates, as indicated by Figure~\ref{fig:accRejRef} plot c). That plot illustrates that the train and test accuracy on the StarLightCurves dataset converges way before trying out all the candidates. However, since the state of the art methods try out all the series segments as candidates, then we also opted for the same approach. In that way, the runtime comparison against the baselines provides an isolated hint on the impact of the pruning strategy.

\subsubsection{An Illustration of The Process}
We present the main idea of our method with the aid of Figure~\ref{fig:accRejRef}. Sub-figures \emph{a), b), c)} display the progress of the method on the StarLightCurves dataset, the largest dataset from the UCR collection~\cite{ucrDatasets}. The fraction of considered (accepted+rejected) shapelets are shown in \emph{a)} with respect to the total candidates in the X-axis. As can be seen, the first few candidates are considered until the accepted and rejected lists are populated with patterns from the dataset. Afterwards, the algorithm starts refusing (pruning/not considering) previously considered candidates within the 25-th percentile threshold, while in the end, an impressive $99.97 \%$ of candidates are pruned. In fact this behavior is not special to the StarLightCurves dataset. We run the algorithm over all the 45 datasets of the UCR collection and measured the fraction of refused candidates as displayed in the histogram of sub-figure \emph{c)}. In average, $99.14 \%$ of candidates can be pruned, with cross-validated values $p,r$ on the training data for each dataset. 

Among the considered candidates, a supervised selection of shapelets is carried on by accepting only those candidates that improve the classification accuracy. Sub-figure \emph{b)} shows that the number of rejections overcomes the number of acceptances as candidates are evaluated, which validates the current belief that very few shapelets can accurately classify a dataset \cite{ye2009time}. As a consequence of the accepted shapelets, the train and test accuracy of the method on the dataset is improved as testified by sub-figure \emph{c)}. With respect to all datasets of the UCR collection, histograms of sub-figures \emph{d), e)} show that in average \textbf{only} $0.06 \%$ of candidates are accepted and $0.81 \%$ are rejected.

\subsubsection{A further intuition}

The similarity based pruning of candidates can be compared to a particular type of clustering where the considered candidates represent centroids. In principle, the mechanism resembles fast online clustering methods \cite{Allan:1998:ONE:290941.290954}. Figure~\ref{fig:clusterIllustration} illustrates how the considered shapelets (blue) can be perceived as an $\epsilon$ threshold clustering of the refused candidates (gray). Each cluster is represented by a hyper-ball of radius $\epsilon$ in a $m$-dimensional space, for $m$ being the shapelet length. For the sake of illustration we selected random points of the shapelets and printed 2-dimensional plots of the 6 considered candidates and 7036 refused candidates from the MALLAT dataset. 

\begin{figure}[!htbp]
\centering
\includegraphics[scale=0.7, trim=2.2cm 16.80cm 8.0cm 6.5cm,clip=true]{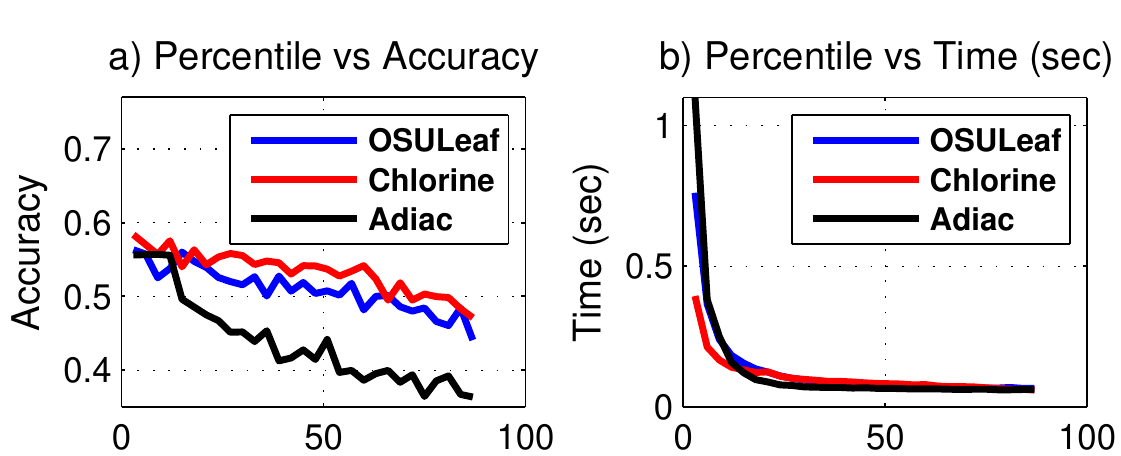}
\caption[]{Impact of alternating the distance threshold's \textbf{percentile} ($p$) value on accuracy, discovery time and the fraction of refused candidates.}
\label{fig:epsilonAnalysis}
\end{figure}

The threshold distance used for pruning similar candidates has a significant effect on the quantity of refused candidates. Figure~\ref{fig:epsilonAnalysis} analyses that the increase of the percentile parameter both deteriorates the classification accuracy (sub-figure \emph{a)}) and significantly shortens the running time (sub-figure \emph{b)}). The higher the distance threshold percentile, the more distant segments will be considered similar and subsequently more candidates will be refused. In order to avoid a severe accuracy deterioration, the percentile parameter $p$ needs to be fixed by cross-validating over the training accuracy.

\subsection{Piecewise Aggregate Approximation (PAA)} 
\label{sec:paa}

The Piecewise Aggregate Approximation (PAA) is a dimensionality reduction technique that shortens time-series by averaging neighbor values \cite{Chakrabarti:2002:LAD:568518.568520}. Algorithm~\ref{alg:paa} illustrates how the time-series of a dataset can be compressed by a ratio $r$. For instance, if $r=\frac{1}{4}$ then every four consecutive points are replaced by their average values.

\begin{algorithm}[!]
 \KwData{Time series data $T \in \R^{N \times M}$, PAA ratio $r \in \left\{ \frac{1}{2},\frac{1}{3},\frac{1}{4}, \dots \right\}$}
 \KwResult{ $ T^\text{PAA} \in \R^{N \times \lceil M \, r \rceil } $ }
 $T \leftarrow \textbf{\bf 0}_{N \times \lceil M \, r \rceil }$\;
 \For{$i \in 1,\dots,N, \, j = 1,\dots,\lceil M \, r \rceil$}{
  \For{$k \in \lceil \frac{1}{r} \left( j-1 \right) + 1  \rceil,\dots, \lceil \frac{j}{r}  \rceil$}{
	$T^\text{PAA}_{i,j} \leftarrow T^\text{PAA}_{i,j} + T_{i,k}$\;
	}
	$T^\text{PAA}_{i,j} \leftarrow T^\text{PAA}_{i,j} \,\, r$\;
 }
 \KwRet{$ T^\text{PAA}$}
 \caption{ \textbf{PiecewiseAggregateApproximation}: Compress every series by a ratio $r$.}
 \label{alg:paa}
\end{algorithm}

\begin{figure*}[]
\centering
\includegraphics[scale=0.7, trim=1.5cm 16.80cm 0.0cm 7.4cm]{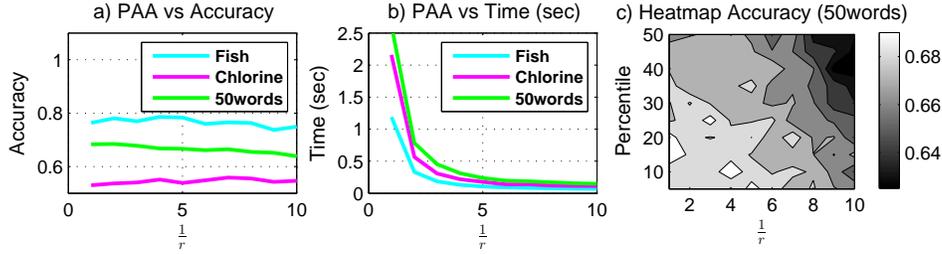}
\caption[]{\textbf{a,b)} Consequence of PAA into accuracy and running time; \textbf{c)} Grid sensitivity of the impact of PAA and the percentile distance threshold over accuracy.}
\label{fig:paaAnalysis} 
\end{figure*}

PAA significantly reduces the discovery time of shapelets as shown in Figure~\ref{fig:paaAnalysis} (sub-figure b) for selected datasets. On the other hand, subfigure a) shows that  the classification accuracy does not deteriorate significantly because time-series data often have a redundancy in length and can be compressed.

The exact amount of PAA reduction and the percentile of the pruning similarity threshold are hyper-parameters that need to be fixed per each dataset using the training data. For instance, Figure~\ref{fig:paaAnalysis} (sub-figure c) illustrates the accuracy heatmap on the 50words dataset as a result of alternating both parameters. As shown, optimal accuracy is achieved for moderate values of percentile threshold and compression. As a contrast, (i) excessive compression and (ii) high threshold percentiles can deteriorate accuracy by (i) destroying informative local patterns by compression and (ii) pruning qualitative variations of shapelet candidates.

\subsection{Algorithmic Analysis of the Runtime Speed-Up}

The running time of shapelet discovery algorithms, which explore candidates among series segments, is upper bounded by the number of candidates in a dataset. Given $N$-many training series of length $M$, the total number of shapelet candidates has an order of $\BigO{N M^2}$, while the time needed to find the best shapelet is $\BigO{N^2 M^4}$. Please note that the discovery time is quadratic in terms of the number of candidates. Applying Piecewise Aggregate Approximation (PAA), in order to reduce the length of time-series by a ratio $r \in \{\frac{1}{2}, \frac{1}{3}, \frac{1}{4}, \dots,... \}$, does alter the runtime complexity into $\BigO{N^2 \left(r \,M\right)^4}$ translated to $\BigO{{\bf r^4} N^2 M^4}$. In other words, PAA reduces the running time by a factor of $\bf r^4$. Furthermore, similarity pruning of candidates has a determinant role in reducing the runtime complexity. Let us denote the fraction of considered candidates as $f := \frac{\# \text{accepted} + \# \text{rejected} }{N M^2}$. Therefore, if executed after a PAA reduction, our algorithm reduces the number of candidates to $\BigO{f N \left( rM \right)^2}$ and impacts the total runtime complexity by $\BigO{f N \left( rM \right)^2 \times \left( N \left( rM \right)^2 + 2N^2 \right) }$, which is upper bounded by $\BigO{f r^4 N^2 M^4}$, since usually $ \left( rM \right)^2 >> 2N $. Ultimately, the expected runtime reduction factor achieved by this paper is upper-bounded by $\bf f r^4$.


There is an addition term that adds up into the runtime complexity: the time needed to check whether any sampled candidate has been previously considered. Such a complexity is $\BigO{N(rM^2) \times f|r \Phi_{*}|}$, in other words, all candidates times the time needed to search for $\epsilon$ similarity on the accepted and rejected lists ($f$-considered candidates having length $|r \Phi_{*}|$). Since  $|r \Phi_{*}| \sim \BigO{rM}$, then the whole operation has a final complexity of $\BigO{fr^3 NM^3}$. Such a complexity is smaller that the time needed to evaluate the accuracy of the candidates ($\BigO{f r^4 N^2 M^4}$), therefore does not alter the big-O complexity. Remember, e.g.: $\BigO{3x^3 + 7x^2 + 100x} \sim \BigO{x^3}$.

Let us illustrate the theoretically expected speed-up via an example. Assume we compress time-series into a quarter of the original lengths, i.e. $r = \frac{1}{2}$. On the other hand, the average fraction of considered shapelets in the UCR datasets is $f = 0.0086$, as previously displayed in Figure~\ref{fig:accRejRef}. Therefore, a run-time reduction factor of $f r^4 = \left(0.0086\right) \left(0.065\right) \approx 5.3 \times 10^{-4}$ is expected. As shown, the expected theoretic runtime speedup can be 4 orders of magnitude compared to the exhaustive shapelet discovery. A detailed analysis of the effects of the dimensionality reduction (PAA compression) and pruning on the runtime performance is provided in Section~\ref{sec:originResults}. Furthermore, in Section~\ref{resultsSec} we will empirically demonstrate that our method is faster than existing shapelet discovery methods. 

\section{Experimental Results}


\subsection{Baselines}

In order to evaluate the efficiency of the proposed method, denoted \textbf{SD}, the fastest state-of-the-art shapelet discovery methods were selected, being:

\begin{enumerate}
\item \textbf{Logical Shapelet \cite{mueen2011logical} (denoted as LS)}: advances the original shapelet discovery method \cite{ye2009time} by one order of magnitude, via: (i) caching and reusing computations, and (ii) applying an admissible pruning of the search space~\cite{mueen2011logical}. 
\item \textbf{Fast Shapelet \cite{rakthanmanon2013fast} (denoted as FS)}: is a recent state-of-the-art method that proposes a random projection technique on the SAX representation by filtering potential candidates~\cite{rakthanmanon2013fast}. FS has been shown to reduce the shapelet discovery time of LS by two to three orders of magnitude~\cite{rakthanmanon2013fast}.
\item \textbf{Improved Fast Shapelet (denoted as FS++)}: is a variation of FS that we created for the sake of being fair to the FS baseline. The original FS paper iterates through all the shapelet lengths from one to the length of the series. In comparison, our method SD iterates through a subset of the possible lengths ($\Phi$) as mentioned in Section~\ref{sec:setup}. In order to be fair (with respect to runtime), we created a variant of the FS, named FS++, that also iterates through the same subsets of shapelet lengths that SD does.
\end{enumerate}

The comparison against the listed state-of-the-art methods will testify the efficiency of our method in terms of runtime scalability. When proposing a faster solution to a supervised learning task, it is crucial to also demonstrate that the speed-up does not deteriorate the prediction accuracy.  For this reason, we payed attention to additionally compare the classification accuracy against the baselines.

\subsection{Setup and Reproducibility}
\label{sec:setup}

\begin{table*}[htbp!]
\scriptsize
\caption{Datasets Statistics (Number of classes (Cls.), Number of series instances (Train/Test) and their length (Len.) )}
\begin{tabular}{|r|l|r|c|r||r|l|r|c|r||r|l|r|c|r|}
\hline
\multicolumn{1}{|c|}{\textbf{No}} & \multicolumn{1}{c|}{\textbf{Dataset}} & \multicolumn{1}{c|}{\textbf{Cls.}} & \multicolumn{1}{c|}{\textbf{Instances}}  & \multicolumn{1}{c||}{\textbf{Len.}} & \multicolumn{1}{c|}{\textbf{No}} & \multicolumn{1}{c|}{\textbf{Dataset}} & \multicolumn{1}{c|}{\textbf{Cls.}} & \multicolumn{1}{c|}{\textbf{Instances}} & \multicolumn{1}{c||}{\textbf{Len.}} & \multicolumn{1}{c|}{\textbf{No}} & \multicolumn{1}{c|}{\textbf{Dataset}} & \multicolumn{1}{c|}{\textbf{Cls.}} & \multicolumn{1}{c|}{\textbf{Instances.}} & \multicolumn{1}{c|}{\textbf{Len.}} \\ \hline \hline
1 & 50words & 50 & 450 / 455 & 270 & 16 & FacesUCR & 14 & 200 / 2050 & 131 & 31 & Sony.I & 2 & 27 / 953 & 65 \\ \hline
2 & Adiac & 37 & 390 / 391 & 176 & 17 & Fish & 7 & 175 / 175 & 463 & 32 & Sony.II & 2 & 20 / 601 & 70 \\ \hline
3 & Beef & 5 & 30 / 30 & 470 & 18 & Gun\_Point & 2 & 50 / 150 & 150 & 33 & StarLight. & 3 & 1000 / 8236 & 1024 \\ \hline
4 & CBF & 3 & 30 / 900 & 128 & 19 & Haptics & 5 & 155 / 308 & 1092 & 34 & SwedishLeaf & 15 & 500 / 625 & 128 \\ \hline
5 & Chlorine. & 3 & 467 / 3840 & 166 & 20 & InlineSkate & 7 & 100 / 550 & 1882 & 35 & Symbols & 6 & 25 / 995 & 398 \\ \hline
6 & CinC\_ECG. & 4 & 40 / 1380 & 1639 & 21 & ItalyPower. & 2 & 67 / 1029 & 24 & 36 & synthetic. & 6 & 300 / 300 & 60 \\ \hline
7 & Coffee & 2 & 28 / 28 & 286 & 22 & Lighting2 & 2 & 60 / 61 & 637 & 37 & Trace & 4 & 100 / 100 & 275 \\ \hline
8 & Cricket\_X & 12 & 390 / 390 & 300 & 23 & Lighting7 & 7 & 70 / 73 & 319 & 38 & Two\_Patterns & 4 & 1000 / 4000 & 128 \\ \hline
9 & Cricket\_Y & 12 & 390 / 390 & 300 & 24 & MALLAT & 8 & 55 / 2345 & 1024 & 39 & TwoLeadECG & 2 & 23 / 1139 & 82 \\ \hline
10 & Cricket\_Z & 12 & 390 / 390 & 300 & 25 & MedicalImages & 10 & 381 / 760 & 99 & 40 & uWave.X & 8 & 896 / 3582 & 315 \\ \hline
11 & Diatom. & 4 & 16 / 306 & 345 & 26 & MoteStrain & 2 & 20 / 1252 & 84 & 41 & uWave.Y & 8 & 896 / 3582 & 315 \\ \hline
12 & ECG200 & 2 & 100 / 100 & 96 & 27 & Non.FatalECG.1 & 42 & 1800 / 1965 & 750 & 42 & uWave.Z & 8 & 896 / 3582 & 315 \\ \hline
13 & ECGFive. & 2 & 23 / 861 & 136 & 28 & Non.FatalECG.2 & 42 & 1800 / 1965 & 750 & 43 & wafer & 2 & 1000 / 6174 & 152 \\ \hline
14 & FaceAll & 14 & 560 / 1690 & 131 & 29 & OliveOil & 4 & 30 / 30 & 570 & 44 & WordsS. & 25 & 267 / 638 & 270 \\ \hline
15 & FaceFour & 4 & 24 / 88 & 350 & 30 & OSULeaf & 6 & 200 / 242 & 427 & 45 & yoga & 2 & 300 / 3000 & 426 \\ \hline
\end{tabular}
\label{tab:dataSetInformation}
\end{table*}

In order to demonstrate the speed-up achievements of the proposed shapelet discovery method, we use the popular collection of time-series datasets from the UCR collection~\cite{ucrDatasets}. For the sake of completeness, we experimented using all the 45 datasets of the collection. The statistics of the datasets are shown in Table~\ref{tab:dataSetInformation}. For each dataset the number of series instances, the number of classes and the length of the time series is presented.

Our Scalable Shapelet Discovery method, denoted as \textbf{SD}, requires the tuning of two parameters, the aggregation ratio $r$ and the threshold percentile $p$. The parameters were searched for each dataset via cross-validation using only the training data. The combination $(r,p)$ that yielded the highest accuracy on train was selected. In case of equal train accuracy scores, then we picked the highest $(r,p)$ values. A grid search was conducted with parameter ranges being $r \in \left\{ \frac{1}{2}, \frac{1}{4}, \frac{1}{8} \right\}$ and $p \in \left\{15,25,35 \right\}$. Finally, the winning combination of parameters was applied over the test data. We would like to note that we used three shapelet lengths for all our experiments, i.e. $L=3$ and $\Phi=\left\{0.2M,0.4M,0.6M \right\}$. In order to promote reproducibility we are presenting the $p,r$ values found by our parameter search in Table~\ref{tab:resultsTab}.

We used the Java programming language to implement our method (SD), while the other baselines (LS, FS, FS++) are implemented in C++. We decided to use the C++ source codes provided and optimized by the respective baseline paper authors \cite{rakthanmanon2013fast,mueen2011logical}, in order to avoid typical allegations on inefficient re-implementations. Finally, we are presenting the exact number of accepted shapelets per each dataset and the respective percentages of the accepted, rejected and refused candidates in the columns merged under "SD Performance". \textbf{All} experiments (both our method and the baselines) were conducted in a \textit{Sun Grid Engine} distributed cluster with 40 node processors, each being \textit{Intel Xeon} E5-2670v2 with speed 2.50GHz and 64GB of shared RAM for all nodes. The operating system was \textit{Linux CentOS} 6.3. All the experiments were launched using the same cluster parameters. 

\emph{The authors are devoted to promote experimental reproducibility. For this reason the source code, all the datasets, the executable file and instructions are provided unconditionally}\footnote{\url{https://www.dropbox.com/sh/btiee2pyn6a989q/AACDfzkkpdYPmgw7pgTgUoeYa}}.

\begin{table*}[htbp]
\center
\caption{ \textbf{Parameters} of SD and \textbf{Results} of SD and SOTA baselines over 45 UCR datasets in terms of shapelets' discovery time and classification accuracy (n/a denotes a 24h time-out)}
\begin{tabular}{|r|l||r|r||r|r|r|r||r|r|r|r||r|r|r|r|}
\hline
\multicolumn{ 1}{|c|}{\textbf{No}} & \multicolumn{ 1}{c||}{\textbf{Dataset}} & \multicolumn{ 2}{c||}{\textbf{SD Parameters}} & \multicolumn{ 4}{c||}{\textbf{SD Acc/Rej/Ref Statistics}} & \multicolumn{ 4}{c||}{\textbf{Discovery Time (seconds)}} & \multicolumn{ 4}{c|}{\textbf{Classification Accuracy}} \\ \cline{ 3- 16}
\multicolumn{ 1}{|c|}{} & \multicolumn{ 1}{c||}{} & \multicolumn{1}{c|}{\textbf{r}} & \multicolumn{1}{c||}{\textbf{p}} & \multicolumn{1}{c|}{\textbf{\#Acc}} & \multicolumn{1}{c|}{\textbf{\%Acc}} & \multicolumn{1}{c|}{\textbf{\%Rej}} & \multicolumn{1}{c||}{\textbf{\%Ref}} & \textbf{LS} & \multicolumn{1}{c|}{\textbf{FS}} & \multicolumn{1}{l|}{\textbf{FS++}} & \multicolumn{1}{c||}{\textbf{SD}} & \multicolumn{1}{c|}{\textbf{LS}} & \multicolumn{1}{c|}{\textbf{FS}} & \multicolumn{1}{l|}{\textbf{FS++}} & \multicolumn{1}{c|}{\textbf{SD}} \\ \hline \hline
1 & 50words & 0.250 & 35 & 39 & 0.0 & 0.1 & 99.8 & n/a & 2198.1 & 35.2 & \bf 0.36 & n/a & 0.511 & 0.446 & \textbf{0.680} \\ \hline
2 & Adiac & 0.500 & 15 & 28 & 0.0 & 0.1 & 99.9 & 12683.2 & 332.6 & 6.4 & \bf 0.25 & \textbf{0.586} & 0.574 & 0.486 & 0.583 \\ \hline
3 & Beef & 0.125 & 35 & 5 & 0.1 & 1.3 & 98.6 & 242.3 & 194.9 & 1.9 & \bf 0.03 & \textbf{0.567} & 0.513 & 0.503 & 0.507 \\ \hline
4 & CBF & 0.500 & 35 & 5 & 0.1 & 0.7 & 99.3 & 66.9 & 10.9 & 0.4 & \bf 0.03 & 0.886 & 0.935 & 0.907 & \textbf{0.975} \\ \hline
5 & Chlorine. & 0.125 & 15 & 13 & 0.1 & 0.2 & 99.7 & 36402.3 & 760.3 & 13.9 & \bf 0.17 & \textbf{0.618} & 0.579 & 0.558 & 0.553 \\ \hline
6 & CinC\_ECG. & 0.125 & 25 & 13 & 0.1 & 1.3 & 98.6 & 2150.0 & 4398.9 & 9.9 & \bf 0.34 & 0.699 & 0.751 & 0.656 & \textbf{0.773} \\ \hline
7 & Coffee & 0.250 & 35 & 4 & 0.1 & 0.3 & 99.6 & 621.9 & 22.5 & 0.2 & \bf 0.03 & \textbf{0.964} & 0.921 & 0.907 & 0.961 \\ \hline
8 & Cricket\_X & 0.250 & 35 & 43 & 0.1 & 0.4 & 99.6 & n/a & 3756.0 & 47.9 & \bf 0.63 & n/a & 0.472 & 0.368 & \textbf{0.672} \\ \hline
9 & Cricket\_Y & 0.250 & 35 & 42 & 0.1 & 0.3 & 99.7 & n/a & 3605.7 & 45.7 & \bf 0.52 & n/a & 0.480 & 0.464 & \textbf{0.675} \\ \hline
10 & Cricket\_Z & 0.250 & 35 & 44 & 0.1 & 0.4 & 99.6 & n/a & 4679.2 & 46.2 & \bf 0.67 & n/a & 0.438 & 0.376 & \textbf{0.673} \\ \hline
11 & Diatom. & 0.125 & 15 & 4 & 0.2 & 1.4 & 98.4 & 184.3 & 15.6 & 0.2 & \bf 0.02 & 0.801 & 0.886 & 0.928 & \textbf{0.896} \\ \hline
12 & ECG200 & 0.125 & 15 & 10 & 0.3 & 2.4 & 97.3 & 618.8 & 16.3 & 0.9 & \bf 0.04 & \textbf{0.870} & 0.766 & 0.786 & 0.818 \\ \hline
13 & ECGFive. & 0.500 & 15 & 5 & 0.1 & 3.2 & 96.7 & 47.6 & 3.6 & 0.1 & \bf 0.03 & 0.994 & \textbf{0.995} & 0.994 & 0.953 \\ \hline
14 & FaceAll & 0.500 & 35 & 40 & 0.0 & 0.5 & 99.5 & 16255.5 & 757.5 & 27.0 & \bf 1.25 & 0.659 & 0.631 & 0.571 & \textbf{0.714} \\ \hline
15 & FaceFour & 0.500 & 35 & 6 & 0.1 & 2.0 & 98.0 & 561.2 & 102.9 & 1.0 & \bf 0.11 & 0.489 & \textbf{0.917} & 0.881 & 0.820 \\ \hline
16 & FacesUCR & 0.500 & 35 & 31 & 0.1 & 1.1 & 98.8 & 2528.5 & 280.3 & 8.7 & \bf 0.33 & 0.662 & 0.703 & 0.654 & \textbf{0.847} \\ \hline
17 & Fish & 0.250 & 25 & 14 & 0.0 & 0.0 & 100.0 & 11153.0 & 935.6 & 6.7 & \bf 0.16 & 0.777 & \textbf{0.809} & 0.785 & 0.755 \\ \hline
18 & Gun\_Point & 0.500 & 25 & 6 & 0.1 & 0.2 & 99.7 & 266.1 & 9.5 & 0.3 & \bf 0.04 & 0.893 & \textbf{0.933} & 0.915 & 0.931 \\ \hline
19 & Haptics & 0.500 & 25 & 13 & 0.0 & 0.0 & 100.0 & n/a & 12491.0 & 31.1 & \bf 1.78 & n/a & \textbf{0.376} & 0.347 & 0.356 \\ \hline
20 & InlineSkate & 0.125 & 15 & 13 & 0.0 & 0.2 & 99.8 & n/a & 22677.2 & 42.6 & \bf 0.61 & n/a & 0.266 & 0.282 & \textbf{0.385} \\ \hline
21 & ItalyPower. & 1.000 & 25 & 6 & 0.1 & 0.9 & 99.0 & 4.9 & 0.4 & 0.1 & \bf 0.02 & \textbf{0.936} & 0.877 & 0.796 & 0.920 \\ \hline
22 & Lighting2 & 0.500 & 35 & 9 & 0.0 & 1.7 & 98.3 & 5297.6 & 1131.3 & 5.0 & \bf 1.89 & 0.426 & 0.707 & 0.698 & \textbf{0.795} \\ \hline
23 & Lighting7 & 0.500 & 35 & 16 & 0.1 & 1.9 & 98.1 & 8619.3 & 322.8 & 3.7 & \bf 0.43 & 0.548 & 0.630 & 0.485 & \textbf{0.652} \\ \hline
24 & MALLAT & 0.125 & 35 & 7 & 0.0 & 0.1 & 99.9 & 1254.9 & 1736.5 & 6.2 & \bf 0.08 & 0.656 & 0.939 & \textbf{0.926} & \textbf{0.926} \\ \hline
25 & MedicalImages & 0.500 & 35 & 34 & 0.1 & 1.1 & 98.9 & 19325.2 & 371.5 & 8.5 & \bf 0.60 & 0.587 & 0.596 & 0.494 & \textbf{0.676} \\ \hline
26 & MoteStrain & 1.000 & 15 & 5 & 0.1 & 8.1 & 91.8 & 6.9 & 3.1 & 0.1 & \bf 0.05 & \textbf{0.832} & 0.783 & 0.767 & 0.783 \\ \hline
27 & Non.FatalECG.1 & 0.250 & 25 & 41 & 0.0 & 0.0 & 100.0 & n/a & 70970.6 & 254.2 & \bf 7.03 & n/a & 0.766 & 0.622 & \textbf{0.814} \\ \hline
28 & Non.FatalECG.2 & 0.125 & 25 & 44 & 0.0 & 0.0 & 100.0 & n/a & 50898.0 & 232.8 & \bf 4.99 & n/a & 0.802 & 0.635 & \textbf{0.855} \\ \hline
29 & OliveOil & 0.125 & 15 & 5 & 0.1 & 0.7 & 99.3 & 502.3 & 107.2 & 0.8 & \bf 0.05 & \textbf{0.833} & 0.723 & 0.773 & 0.790 \\ \hline
30 & OSULeaf & 0.125 & 25 & 21 & 0.1 & 0.2 & 99.7 & 14186.5 & 1629.7 & 20.0 & \bf 0.15 & \textbf{0.686} & 0.680 & 0.555 & 0.566 \\ \hline
31 & Sony.I & 1.000 & 35 & 4 & 0.1 & 0.8 & 99.1 & 4.6 & 1.1 & 0.1 & \bf 0.02 & \textbf{0.860} & 0.686 & 0.802 & 0.850 \\ \hline
32 & Sony.II & 1.000 & 35 & 5 & 0.1 & 1.6 & 98.3 & 9.8 & 1.3 & 0.1 & \bf 0.03 & 0.846 & 0.792 & \textbf{0.945} & 0.780 \\ \hline
33 & StarLight. & 0.125 & 25 & 20 & 0.0 & 0.0 & 100.0 & n/a & 21473.5 & 78.5 & \bf 3.19 & n/a & \textbf{0.942} & 0.932 & 0.933 \\ \hline
34 & SwedishLeaf & 0.500 & 25 & 30 & 0.0 & 0.1 & 99.9 & 11953.6 & 451.7 & 12.9 & \bf 0.36 & 0.813 & 0.779 & 0.725 & \textbf{0.849} \\ \hline
35 & Symbols & 0.250 & 25 & 4 & 0.1 & 1.0 & 98.9 & 894.3 & 93.0 & 0.6 & \bf 0.04 & 0.643 & 0.933 & 0.756 & \textbf{0.865} \\ \hline
36 & synthetic. & 0.250 & 35 & 11 & 0.1 & 0.2 & 99.7 & 3667.4 & 63.9 & 3.6 & \bf 0.07 & 0.470 & 0.922 & 0.870 & \textbf{0.983} \\ \hline
37 & Trace & 0.500 & 35 & 7 & 0.0 & 0.1 & 99.8 & 4626.9 & 181.0 & 1.7 & \bf 0.13 & \textbf{1.000} & 0.994 & 0.999 & 0.965 \\ \hline
38 & Two\_Patterns & 0.500 & 35 & 38 & 0.0 & 0.1 & 99.9 & 65783.1 & 957.2 & 37.7 & \bf 1.71 & 0.539 & 0.310 & 0.753 & \textbf{0.981} \\ \hline
39 & TwoLeadECG & 1.000 & 25 & 4 & 0.1 & 0.4 & 99.6 & 14.3 & 1.3 & 0.0 & \bf 0.02 & 0.856 & \textbf{0.928} & 0.798 & 0.867 \\ \hline
40 & uWave.X & 0.250 & 25 & 44 & 0.0 & 0.3 & 99.7 & n/a & 4827.5 & 54.1 & \bf 4.94 & n/a & 0.707 & 0.580 & \textbf{0.761} \\ \hline
41 & uWave.Y & 0.250 & 25 & 41 & 0.0 & 0.2 & 99.8 & n/a & 4379.6 & 56.6 & \bf 3.69 & n/a & 0.608 & 0.466 & \textbf{0.671} \\ \hline
42 & uWave.Z & 0.125 & 25 & 37 & 0.0 & 0.2 & 99.8 & n/a & 5215.9 & 50.9 & \bf 1.83 & n/a & 0.627 & 0.565 & \textbf{0.676} \\ \hline
43 & wafer & 0.500 & 35 & 10 & 0.0 & 0.1 & 99.9 & 34653.1 & 190.5 & 5.0 & \bf 1.39 & \textbf{0.999} & 0.998 & 0.949 & 0.993 \\ \hline
44 & WordsS. & 0.250 & 25 & 35 & 0.1 & 0.4 & 99.5 & n/a & 1140.0 & 18.7 & \bf 0.31 & n/a & 0.437 & 0.389 & \textbf{0.625} \\ \hline
45 & yoga & 0.250 & 15 & 17 & 0.0 & 0.1 & 99.9 & 11389.0 & 1711.6 & 11.2 & \bf 0.34 & \textbf{0.740} & 0.705 & 0.697 & 0.625 \\ \hline \hline
\multicolumn{8}{|c||}{ \textbf{Total Wins} } & \bf 0 & \bf 0 & \bf 0 & \bf 45 & \bf 13 & \bf 9 & \bf 2 & \bf 21 \\ \hline 
\multicolumn{8}{|c||}{ \textbf{Average Rank} } & \bf 0.000 & \bf 0.000 & \bf 0.000 & \bf 1.000 & \bf 2.313 & \bf 2.178 & \bf 3.089 & \bf 1.889 \\ \hline
\end{tabular}
\label{tab:resultsTab}
\end{table*}

\subsection{Highly Qualitative Runtime Results}
\label{resultsSec}

\begin{figure}[h]
\centering
\includegraphics[scale=0.5105, trim=2cm 7.09cm 0.0cm 6.0cm]{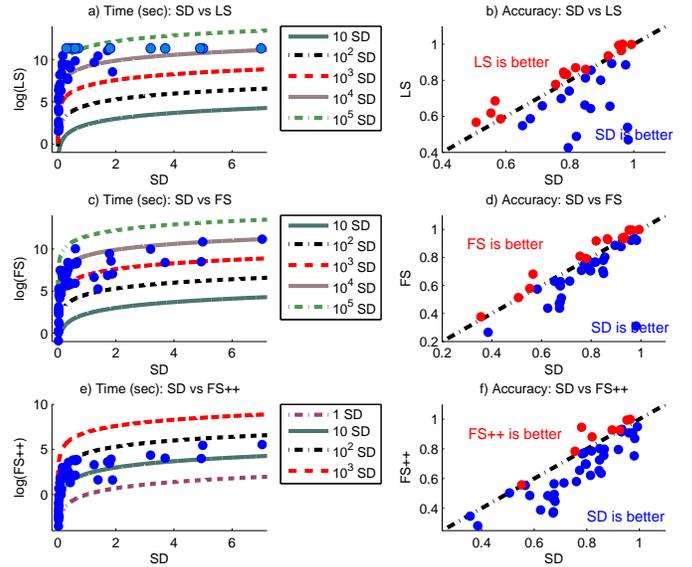}
\caption[]{ Time and accuracy comparison of our method (denoted \textbf{SD}) against state-of-the-art methods both in terms of discovery time and classification accuracy for all the 45 UCR datasets.}
\label{fig:speedUpAnalysis}
\end{figure}

The empirical results include both the discovery time and the classification accuracy of our method SD against baselines for 45 UCR datasets. Table~\ref{tab:resultsTab} contains a list of results per dataset, where the discovery time is measured in seconds. A time-out threshold of 24 hours was set for the discovery of shapelets of a single dataset. As can be seen, the Logical Shapelet (LS) exceeded the time-out threshold in a considerable number of datasets. The reader is invited to notice that 24 hours (86400 seconds) is a very large threshold, given that our method SD often finds the shapelets within a fraction of one second, as for instance in the 50words dataset. Finally, we are presenting the exact number of accepted shapelets per each dataset and the respective percentages of the accepted, rejected and refused candidates in the columns merged under "SD Performance". 

It can be clearly deduced that our method SD is faster than the fastest existing baselines LS \cite{mueen2011logical} and FS \cite{rakthanmanon2013fast}. There is no dataset where any of the baselines is faster. Even, our modification of FS, i.e. the FS++, is considerably slower than SD. For instance, it took only 3.19 seconds for our method to find the shapelets of the StarLightCurves dataset, which has 1000 training instances each having 1024 points. The high-level conclusion from the discovery time results is: "Since the introduction of shapelets in 2009, time-series community believed shapelets are very useful classification patterns, but finding them is slow. This paper demonstrates that shapelets can be discovered very fast." 

The discovery time measurements do not include the time needed by a practitioner to tune the parameters of the methods. While our method has two parameters ($p$ and $r$, totaling $3\times 3=9$ combinations, see Section~\ref{sec:setup}), the strongest baseline (Fast Shapelet) has more parameters, concretely four: the reduced dimensionality and cardinality of SAX, the random projection iterations and the number of SAX candidates (denoted d,c,r,k in the original paper~\cite{rakthanmanon2013fast}).

\subsection{Competitive Prediction Accuracy}

Yet, our results are atypical in another positive aspect. Most scalability papers propose speed-ups of the learning time by sacrificing a certain fraction of the prediction accuracy. In contrast, our results show that our method is both faster and more accurate than the baselines. The winning method that achieves the highest accuracy on each dataset (on each row) is distinguished in bold. Our method has more wins than the baselines (21 wins against 13 of the second best method) and also a better rank (1.889 against 2.178 of the second best method). The accuracy improvement arises from the joint interaction of accepted shapelets as predictors (distance matrix X in Algorithm~\ref{alg:discoverShapelets}), while the baselines measure the quality of each shapelet separately, without considering their interactions~\cite{ye2009time,mueen2011logical,rakthanmanon2013fast}. Incorporating the interactions among shapelets into the prediction model has been recently shown to achieve high classification accuracy \cite{grabocka2014kdd}.

\begin{table*}[htbp]
\centering
\caption{Modular Decomposition of The Performance of Our Method (SD), N/A Denotes a 24H time-out}
\begin{tabular}{|r|l||r|r|r|r||c|c|c|c|}
\hline
\multicolumn{1}{|c|}{\textbf{}} & \multicolumn{1}{c||}{\textbf{}} & \multicolumn{ 4}{c||}{\textbf{Discovery Time (seconds)}} & \multicolumn{ 4}{c|}{\textbf{Classification Accuracy}} \\ 
\cline{ 3- 10}
\multicolumn{ 1}{|c|}{\textbf{No}} & \multicolumn{ 1}{c||}{\textbf{Dataset}} & \multicolumn{1}{l|}{\textbf{\xmark \, PAA}} & \multicolumn{1}{l|}{\textbf{\cmark \, PAA}} & \multicolumn{1}{l|}{\textbf{\xmark \, PAA}} & \multicolumn{1}{l||}{\textbf{\cmark \, PAA}} & \multicolumn{1}{l|}{\textbf{\xmark \, PAA}} & \multicolumn{1}{l|}{\textbf{\cmark \, PAA}} & \multicolumn{1}{l|}{\textbf{\xmark \, PAA}} & \multicolumn{1}{l|}{\textbf{\cmark \, PAA}} \\ \cline{ 3- 10}
\multicolumn{ 1}{|l|}{} & \multicolumn{ 1}{l||}{} & \multicolumn{1}{l|}{\textbf{\xmark \, prun.}} & \multicolumn{1}{l|}{\textbf{\xmark \, prun.}} & \multicolumn{1}{l|}{\textbf{\cmark \, prun.}} & \multicolumn{1}{l||}{\textbf{\cmark \, prun.}} & \multicolumn{1}{l|}{\textbf{\xmark \, prun.}} & \multicolumn{1}{l|}{\textbf{\xmark \, prun.}} & \multicolumn{1}{l|}{\textbf{\cmark \, prun.}} & \multicolumn{1}{l|}{\textbf{\cmark \, prun.}} \\ \hline \hline
1 & 50words & 4028.85 & 154.74 & 5.24 & 0.36 & 0.684 & \bf 0.701 & 0.679 & 0.680 \\ \hline
2 & Adiac & 799.06 & 153.21 & 0.89 & 0.25 & \bf 0.624 & 0.555 & 0.604 & 0.583 \\ \hline
3 & Beef & 61.35 & 0.65 & 0.54 & 0.03 & 0.533 & \bf 0.600 & 0.500 & 0.507 \\ \hline
4 & CBF & 2.22 & 0.57 & 0.37 & 0.03 & \bf 0.992 & 0.964 & 0.929 & 0.975 \\ \hline
5 & Chlorine. & 1598.05 & 30.14 & 2.40 & 0.17 & 0.527 & \bf 0.596 & 0.539 & 0.553 \\ \hline
6 & CinC\_ECG. & 3718.48 & 11.74 & 12.71 & 0.34 & \bf 0.809 & 0.768 & 0.776 & 0.773 \\ \hline
7 & Coffee & 11.79 & 1.27 & 0.39 & 0.03 & \bf 0.964 & 0.893 & 0.893 & 0.961 \\ \hline
8 & Cricket\_X & 4218.58 & 141.80 & 23.63 & 0.63 & \bf 0.697 & \bf 0.697 & 0.669 & 0.672 \\ \hline
9 & Cricket\_Y & 3953.86 & 137.75 & 14.20 & 0.52 & \bf 0.715 & 0.687 & 0.677 & 0.675 \\ \hline
10 & Cricket\_Z & 5313.96 & 132.06 & 40.17 & 0.67 & 0.700 & 0.682 & \bf 0.726 & 0.673 \\ \hline
11 & Diatom. & 5.00 & 0.50 & 0.50 & 0.02 & 0.915 & \bf 0.948 & 0.827 & 0.896 \\ \hline
12 & ECG200 & 14.59 & 0.70 & 0.42 & 0.04 & 0.820 & 0.800 & \bf 0.830 & 0.818 \\ \hline
13 & ECGFiveDays & 1.41 & 0.40 & 0.36 & 0.03 & \bf 0.999 & 0.945 & 0.981 & 0.953 \\ \hline
14 & FaceAll & 1276.24 & 297.23 & 4.87 & 1.25 & 0.720 & \bf 0.731 & 0.724 & 0.714 \\ \hline
15 & FaceFour & 18.04 & 3.10 & 1.21 & 0.11 & 0.852 & 0.898 & \bf 0.943 & 0.820 \\ \hline
16 & FacesUCR & 107.35 & 24.74 & 2.70 & 0.33 & \bf 0.871 & 0.868 & 0.841 & 0.847 \\ \hline
17 & Fish & 1808.85 & 46.46 & 1.61 & 0.16 & 0.817 & \bf 0.846 & 0.800 & 0.755 \\ \hline
18 & Gun\_Point & 7.69 & 1.55 & 0.60 & 0.04 & 0.900 & 0.913 & \bf 0.953 & 0.931 \\ \hline
19 & Haptics & 17273.44 & 2634.99 & 6.59 & 1.78 & 0.354 & \bf 0.373 & 0.321 & 0.356 \\ \hline
20 & InlineSkate & 34776.14 & 99.61 & 19.82 & 0.61 & \bf 0.411 & 0.342 & 0.313 & 0.385 \\ \hline
21 & ItalyPower. & 0.77 & 0.49 & 0.62 & 0.02 & \bf 0.936 & 0.925 & 0.915 & 0.920 \\ \hline
22 & Lighting2 & 843.42 & 90.84 & 12.23 & 1.89 & \bf 0.852 & 0.836 & 0.836 & 0.795 \\ \hline
23 & Lighting7 & 120.39 & 20.15 & 4.89 & 0.43 & 0.699 & \bf 0.740 & 0.685 & 0.652 \\ \hline
24 & MALLAT & 2295.97 & 6.25 & 1.99 & 0.08 & 0.909 & 0.938 & \bf 0.941 & 0.926 \\ \hline
25 & MedicalImages & 349.15 & 57.75 & 1.76 & 0.60 & 0.625 & 0.658 & 0.668 & \bf 0.676 \\ \hline
26 & MoteStrain & 0.91 & 0.62 & 0.21 & 0.05 & 0.734 & \bf 0.815 & 0.777 & 0.783 \\ \hline
27 & Non.FatalECG.1 & \multicolumn{1}{r|}{n/a} & 35833.59 & 36.79 & 7.03 & \multicolumn{1}{c|}{n/a} & \bf 0.840 & 0.795 & 0.814 \\ \hline
28 & Non.FatalECG.2 & \multicolumn{1}{r|}{n/a} & 11086.13 & 58.18 & 4.99 & \multicolumn{1}{c|}{n/a} & 0.852 & \bf 0.858 & 0.855 \\ \hline
29 & OliveOil & 75.17 & 0.90 & 1.12 & 0.05 & \bf 0.900 & 0.800 & 0.700 & 0.790 \\ \hline
30 & OSULeaf & 2379.27 & 16.64 & 3.18 & 0.15 & 0.570 & 0.541 & \bf 0.583 & 0.566 \\ \hline
31 & Sony.I & 1.28 & 0.62 & 0.76 & 0.02 & 0.829 & \bf 0.902 & 0.792 & 0.850 \\ \hline
32 & Sony.II & 0.46 & 0.47 & 0.86 & 0.03 & 0.727 & 0.774 & 0.742 & \bf 0.780 \\ \hline
33 & StarLightCurves & \multicolumn{1}{r|}{n/a} & 4673.16 & 74.14 & 3.19 & \multicolumn{1}{c|}{n/a} & \bf 0.933 & 0.929 & \bf 0.933 \\ \hline
34 & SwedishLeaf & 830.60 & 301.63 & 1.24 & 0.36 & \bf 0.869 & 0.856 & 0.856 & 0.849 \\ \hline
35 & Symbols & 27.58 & 1.64 & 0.58 & 0.04 & 0.805 & 0.787 & 0.819 & \bf 0.865 \\ \hline
36 & synthetic. & 51.03 & 6.01 & 0.56 & 0.07 & 0.980 & \bf 0.993 & 0.980 & 0.983 \\ \hline
37 & Trace & 138.09 & 25.15 & 0.60 & 0.13 & 0.950 & \bf 0.990 & 0.960 & 0.965 \\ \hline
38 & Two\_Patterns & 4572.63 & 1216.45 & 2.78 & 1.71 & 0.985 & 0.984 & \bf 0.986 & 0.981 \\ \hline
39 & TwoLeadECG & 0.54 & 0.88 & 0.41 & 0.02 & \bf 0.932 & 0.774 & \bf 0.932 & 0.867 \\ \hline
40 & uWave.X & 27142.53 & 1565.73 & 19.46 & 4.94 & 0.757 & 0.745 & \bf 0.762 & 0.761 \\ \hline
41 & uWave.Y & 25276.28 & 1385.23 & 16.74 & 3.69 & 0.647 & 0.643 & \bf 0.671 & \bf 0.671 \\ \hline
42 & uWave.Z & 24532.05 & 513.11 & 14.09 & 1.83 & 0.662 & 0.668 & \bf 0.681 & 0.676 \\ \hline
43 & wafer & 6352.87 & 1750.96 & 3.31 & 1.39 & 0.994 & 0.994 & \bf 0.995 & 0.993 \\ \hline
44 & WordsS. & 1220.31 & 44.25 & 3.13 & 0.31 & 0.627 & \bf 0.639 & 0.607 & 0.625 \\ \hline
45 & yoga & 5098.73 & 254.54 & 3.05 & 0.34 & \bf 0.812 & 0.802 & 0.799 & 0.625 \\ \hline \hline
\multicolumn{6}{|c||}{\bf Absolute Wins} & \bf 14.0 & \bf 15.0 & \bf 12.0 & \bf 4.0 \\ \hline
\multicolumn{6}{|c||}{\bf Ranks} & $\bf 2.2 \pm 1.1$ & $\bf 2.3 \pm 1.1$ & $\bf 2.5 \pm 1.2$ & $\bf 2.7 \pm 0.9$ \\ \hline
\end{tabular}
\label{tab:originsResults}
\end{table*}

\subsection{Speed-Up Analysis}

In order to show the speed-up factor of our method with respect to the (former) state-of-the-art, we provide another presentation of the results in Figure~\ref{fig:speedUpAnalysis}. The three plots on the left side show the discovery time of SD in x-axis and the logarithm of the discovery time of each baseline as the y-axis. As can be easily observed from the illustrative order lines, SD is 4 to 5 orders of magnitude faster than the Logical Shapelet (LS) and 3 to 4 orders of magnitude faster than the Fast Shapelet (FS). The datasets where LS exceeds the 24 hour threshold are depicted in light blue. In addition, FS++ is faster than FS because it iterates over less shapelet length sizes, yet it is still 1 to 2 orders of magnitude slower than SD. 

The plots on the right represent scatter plots of the classification accuracy of SD against the baselines. While generally better than LS and FS, our method SD is largely superior to FS++. Such a finding indicates that the accuracy of the Fast Shapelet (FS) is dependent on trying shapelet candidates from a fine-grained set of lengths, while our method is very accurate even though it iterates over few shapelet lengths. 

\subsection{A Modular Decomposition of the Performance}
\label{sec:originResults}

We have already seen that our proposed method, SD, outperforms significantly the state-of-the-art in terms of runtime and produces even better prediction accuracy. Nevertheless, there are a couple of questions that can be addressed to our method, such as:

\begin{enumerate}
\item{What fraction of SD's runtime reduction is attributed to the novel candidate pruning and what fraction to the PAA compression?}
\item{To what extent does pruning deteriorate the prediction accuracy?}
\end{enumerate}

In order to address those analytic questions we will decompose our method in a modular fashion. Our method, SD, conducts both a PAA approximation and a pruning by the parameters $r,p$ provided in Table~\ref{tab:resultsTab}. In order to isolate the effect of compression and pruning we are creating four variants of our method, namely all the permutations "With/Without PAA compression" and "With/Without Pruning" (w.r.t. to $p,r$ from Table~\ref{tab:resultsTab}). All the decomposed results of the SD variants are shown in Table~\ref{tab:originsResults}. Note that "No pruning" means $p=0$, while "no PAA" means $r=1$. The variant with both pruning and PAA is the same as SD from Section~\ref{resultsSec}, which already was shown to be superior to the state of the art. 

\begin{figure}[t]
\centering
\includegraphics[scale=0.5105, trim=2cm 14.3cm 0.0cm 6.0cm]{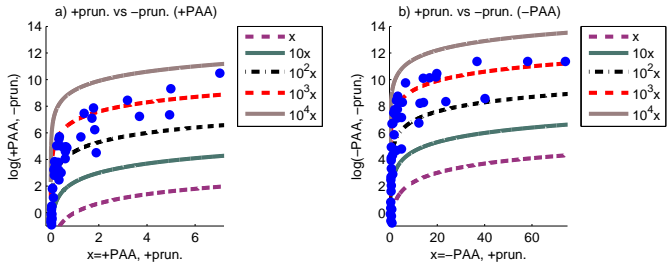}
\caption[]{Runtime comparison (seconds) plots among variants of SD with and without pruning}
\label{fig:originPrunNpPrun}
\vspace{-0.3cm}
\end{figure}

Looking into the results of Table~\ref{tab:originsResults}, it is important to observe that the variant with PAA compression alone is significantly faster than the variant without compression (columns 4 vs column 3). However, using pruning without compression is much faster than the exhaustive approach and also much faster than compression alone (column 5 vs. columns 3,4). When pruning and compression are combined (column 6), then the runtime reduction effect multiplies. More concretely, Figure~\ref{fig:originPrunNpPrun} analyses the runtime reduction of SD variants: that use pruning (X-axis) against variants without pruning (Y-axis) for both scenarios with PAA (plot a)) or without PAA (plot b)) compression. As can be clearly deduced, pruning alone has a significant effect on the runtime reduction by 3 to 4 orders of magnitude, compared to the cases where no pruning is employed. While PAA helps our method to be even faster, it is clear that the lion share of the speedup arises from the proposed pruning mechanism.

There is still a concern on how does pruning affect the classification accuracy. The prediction accuracy results are demonstrated in Table~\ref{tab:originsResults} for all the datasets, with the winning variant emphasized in bold. The total wins and the ranks of the variants indicate that the best prediction performance is attributed to the exhaustive methods (no pruning, columns 7,8). Such a finding is natural because exhaustive approaches consider all the candidate variants and can extracts more qualitative minimum distance features. Yet, are the results of the exhaustive variants better with a \textbf{statistical significance} margin? Table~\ref{tab:wilcoxonStatSign} illustrates the p-values of a Wilcoxon Signed Rank test of statistical significance, for a two-tailed hypothesis with a significance level of 5\% ($\alpha=0.05$).

\begin{table}[h]
\centering
\caption{Wilcoxon Statistical Significance Test: p-values (Significance Level 5\%, Two-Tailed Hypothesis)}
\begin{tabular}{|l||c|c|c|c|}
\hline
\multicolumn{1}{|r||}{} & \multicolumn{1}{c|}{\xmark \, PAA} & \multicolumn{1}{c|}{\cmark \, PAA} & \multicolumn{1}{c|}{\xmark \, PAA} & \multicolumn{1}{c|}{\cmark \, PAA} \\ 
\multicolumn{1}{|r||}{} & \multicolumn{1}{c|}{\xmark \, prun.} & \multicolumn{1}{c|}{\xmark \, prun.} & \multicolumn{1}{c|}{\cmark \, prun.} & \multicolumn{1}{c|}{\cmark \, prun.} \\
\hline \hline
\multicolumn{1}{|r||}{\xmark \, PAA, \xmark \, prun.} & - & 0.904 & \bf 0.119 & 0.046 \\ \hline
\multicolumn{1}{|r||}{\cmark \, PAA, \xmark \, prun.} & 0.904 & - & 0.153 & \bf 0.112 \\ \hline
\multicolumn{1}{|r||}{\xmark \, PAA, \cmark \, prun.} & \bf 0.119 & 0.153 & - & 0.873 \\ \hline
\multicolumn{1}{|r||}{\cmark \, PAA, \cmark \, prun.} & 0.046 & \bf 0.112 & 0.873 & - \\ \hline
\end{tabular}
\label{tab:wilcoxonStatSign}
\end{table}

The p-values which compare variants that use pruning against variants that does not use pruning are shown in bold and correspond to $p=0.119, p=0.112$. Therefore, the prediction quality using pruning is not significantly (significance means $p < 0.05$) worse than the exhaustive approach. The final message of this section is: "Pruning of candidates provides 3 to 4 orders of runtime speedup without any statistically significant deterioration in terms of classification accuracy.".

\section{Conclusion}

Shapelets represent discriminative segments of a time-series dataset and the distances of time-series to shapelets are shown to be successful features for classification. The discovery of shapelets is currently conducted by trying out candidates from the segments (sub-sequences) of the time-series. Since the number of candidate segments is large, the time-series community has spent efforts on speeding up the discovery time of shapelets. This paper proposed a novel method that prunes the candidates based on a distance threshold to previously considered other similar candidates. In a parallel fashion, a novel supervised selection filters those shapelets that boost classification accuracy. We empirically showed that our method is 3-4 orders of magnitude faster than the fastest existing shapelet discovery methods, while providing a better prediction accuracy.

%



\vspace{-0.2cm}
\bibliographystyle{IEEEtran}
\bibliography{grabocka2014g-icdm}
%

\end{document}